\documentclass[10pt,twocolumn,letterpaper]{article}

\usepackage[final]{cvpr}      

\usepackage{placeins}
\usepackage{graphicx}
\usepackage{amsmath}
\usepackage{amssymb}
\usepackage{booktabs}
\usepackage{float}

\usepackage[pagebackref,breaklinks,colorlinks]{hyperref}
\usepackage[capitalize]{cleveref}
\crefname{section}{Sec.}{Secs.}
\Crefname{section}{Section}{Sections}
\Crefname{table}{Table}{Tables}
\crefname{table}{Tab.}{Tabs.}

\usepackage[pagebackref,breaklinks,colorlinks]{hyperref}

\usepackage[capitalize]{cleveref}
\crefname{section}{Sec.}{Secs.}
\Crefname{section}{Section}{Sections}
\Crefname{table}{Table}{Tables}
\crefname{table}{Tab.}{Tabs.}

\begin{document}

\title{Reconstructing Tornadoes in 3D with Gaussian Splatting}

\author{Adam Yang \\ University of Maryland \\ \href{mailto:ayang115@umd.edu}{ayang115@umd.edu}
\and
Nadula Kadawedduwa\\ University of Maryland \\ \href{mailto:nkadawed@umd.edu}{nkadawed@umd.edu}
\and
Tianfu Wang \\ University of Maryland
\and
Sunny Sharma \\ University of Maryland
\and
Emily F. Wisinski \\ University of Maryland
\and
Jhayron S. Pérez-Carrasquilla \\ University of Maryland
\and
Kyle J. C. Hall \\ University of Maryland
\and
Dean Calhoun \\ University of Maryland
\and
Jonathan Starfeldt \\ University of Maryland
\and
Timothy P. Canty \\ University of Maryland
\and
Maria J. Molina \\ University of Maryland
\and
Christopher A.~Metzler \\ University of Maryland
\\
}
\maketitle

\begin{abstract}
    Accurately reconstructing the 3D structure of tornadoes could significantly enhance our understanding of this highly destructive weather phenomenon, ultimately assisting in adequate preparedness and timely emergency management response. 
    While modern 3D scene reconstruction techniques, such as 3D Gaussian splatting (3DGS), could provide a valuable tool for reconstructing the 3D structure of tornadoes, at present, we are critically lacking a controlled tornado dataset with which to develop and validate these tools. 
    In this work, we capture and release a novel multiview dataset of a small lab-based tornado. 
    We demonstrate that one can effectively reconstruct and visualize the 3D structure of this tornado using 3DGS. 
    \\
    Project Page: \href{https://adamy03.github.io/Reconstruction-Tornadoes-with-Neural-Inverse-Rendering/}{Reconstructing Tornadoes with 3DGS}
    \\
    Dataset: \href{https://umd.box.com/s/xcosiygyau5bk2cxzpm92zw30jmtwx82}{Multi-view tornado chamber capture}

\end{abstract}

\section{Introduction}
The first coordinated, large-scale tornado field study was VORTEX-1 (Verification of the Origins of Rotation in Tornadoes Experiment) \cite{rasmussen1994verification}, which took place in 1994-1995 over the Great Plains of the United States (US). More recent field campaigns have been focused on the Mid-South and Southeast US \cite{kosiba2024propagation}, where terrain, diurnal cycle, convective mode, and population vulnerability amplify tornado impact risk \cite{strader2024changes}.  In these tornado field campaigns, multi-Doppler radar analyses were attempted to reconstruct the 3D wind field of tornadoes \cite{wurman1997design}, which is challenging due to safety constraints surrounding the precise placement geometry needed during rapid tornado evolution. While these field campaigns gathered landmark observations on tornadoes, including at least one well-documented case of `true' multi-Doppler sampling \cite{kosiba2013three}, understanding of tornado formation and evolution remains limited, partly due to the difficulty in measuring the 3D wind field of tornadoes. Proliferation of video surveillance and `storm chaser' footage is a large source of up-close tornado imagery \cite{seimon2016crowdsourcing} that could be leveraged to further our 3D understanding of tornadoes.

Novel view synthesis (NVS) is a canonical task in computer vision and machine learning. Advances in NVS and deep learning have led to vastly improved techniques for generating 3D models from multi-view camera scenes or videos \cite{mildenhall2020nerfrepresentingscenesneural} \cite{kerbl3Dgaussians}. However, the application of such models to atmospheric phenomena, such as tornadoes, remains understudied. To address this, our work seeks to bridge advances in NVS with 3D tornado imaging and modeling. 

What follows outlines our application of 3D Gaussian Splatting (3DGS), as demonstrated in \cite{kerbl3Dgaussians}, to collected footage of a simulated tornado. We will discuss the experimental setup, 3D modeling framework, and postprocessing steps.  Following this, we discuss our results using the proposed pipeline and propose future directions. A summary of our contributions is as follows:
\begin{itemize}
    \item We first introduce a novel dataset of a simulated tornado.
    \item Next, evaluate the performance of vanilla 3DGS on our collected footage.
    \item Lastly, we propose further work to improve our proposed pipeline.
\end{itemize}

\begin{figure}[h!]
    \centering
    \includegraphics[width=1\linewidth]{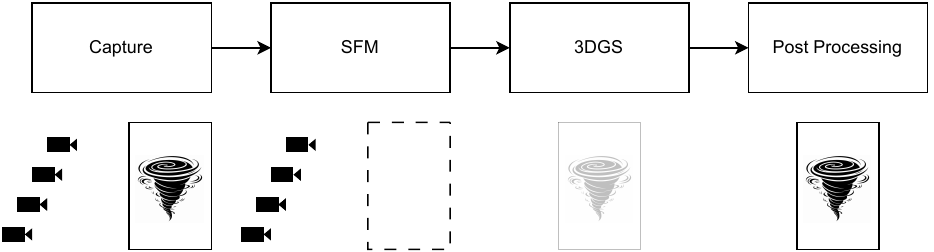}
    \caption{\textbf{Reconstruction pipeline.} Procedure starts with idealized tornado capture, with select frames used for reconstruction. Structure from motion (SfM) is used to get an initial point cloud, which is then used to initialize 3DGS. Lastly, using the resulting point cloud, post-processing removes incorrect Gaussians for the final render.}
    \label{fig:workflow}
\end{figure}

\begin{figure*}[h!]
    \centering
    \begin{tabular}{ccccc} 
      \includegraphics[width=0.15\textwidth]{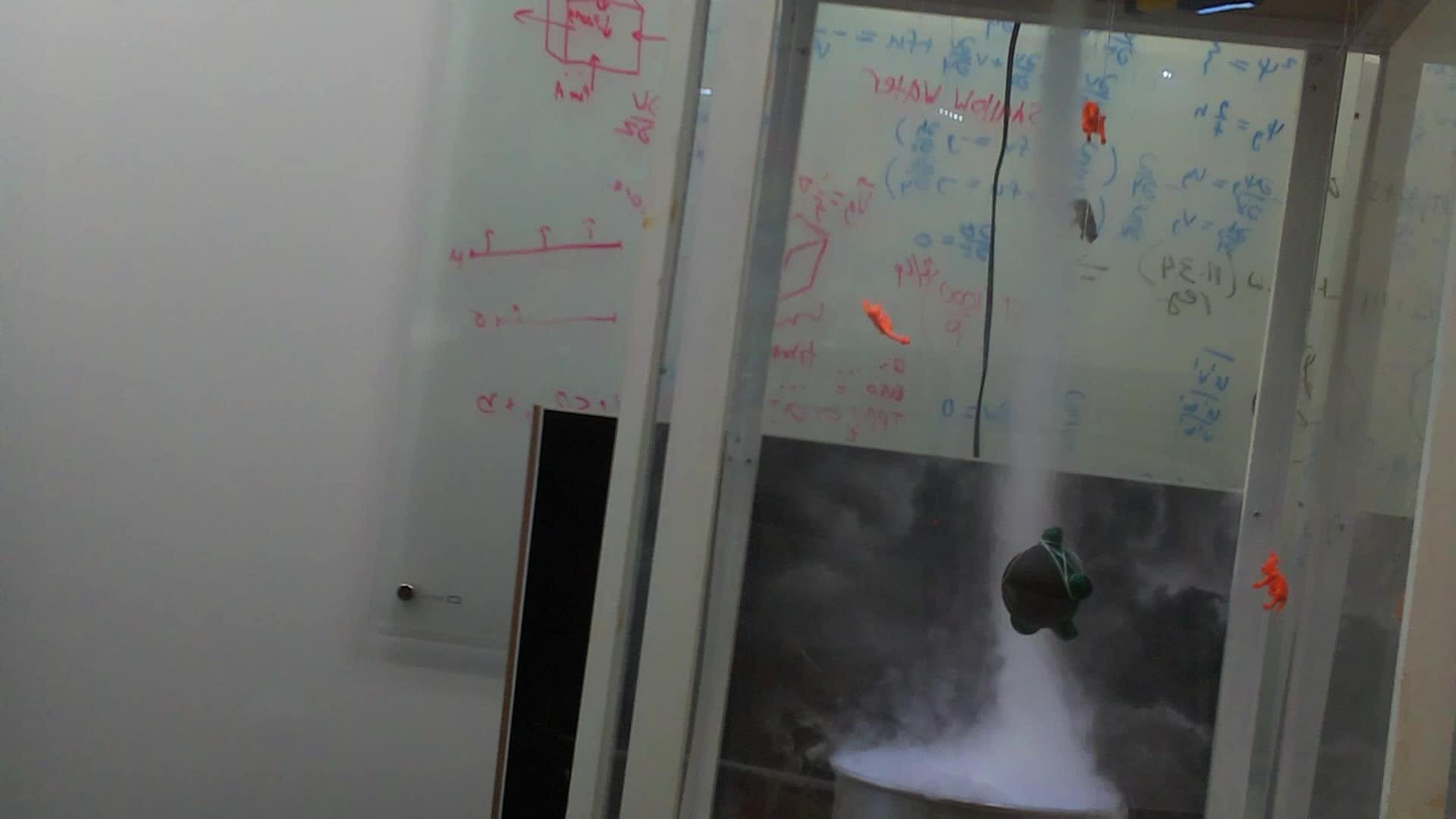} &
      \includegraphics[width=0.15\textwidth]{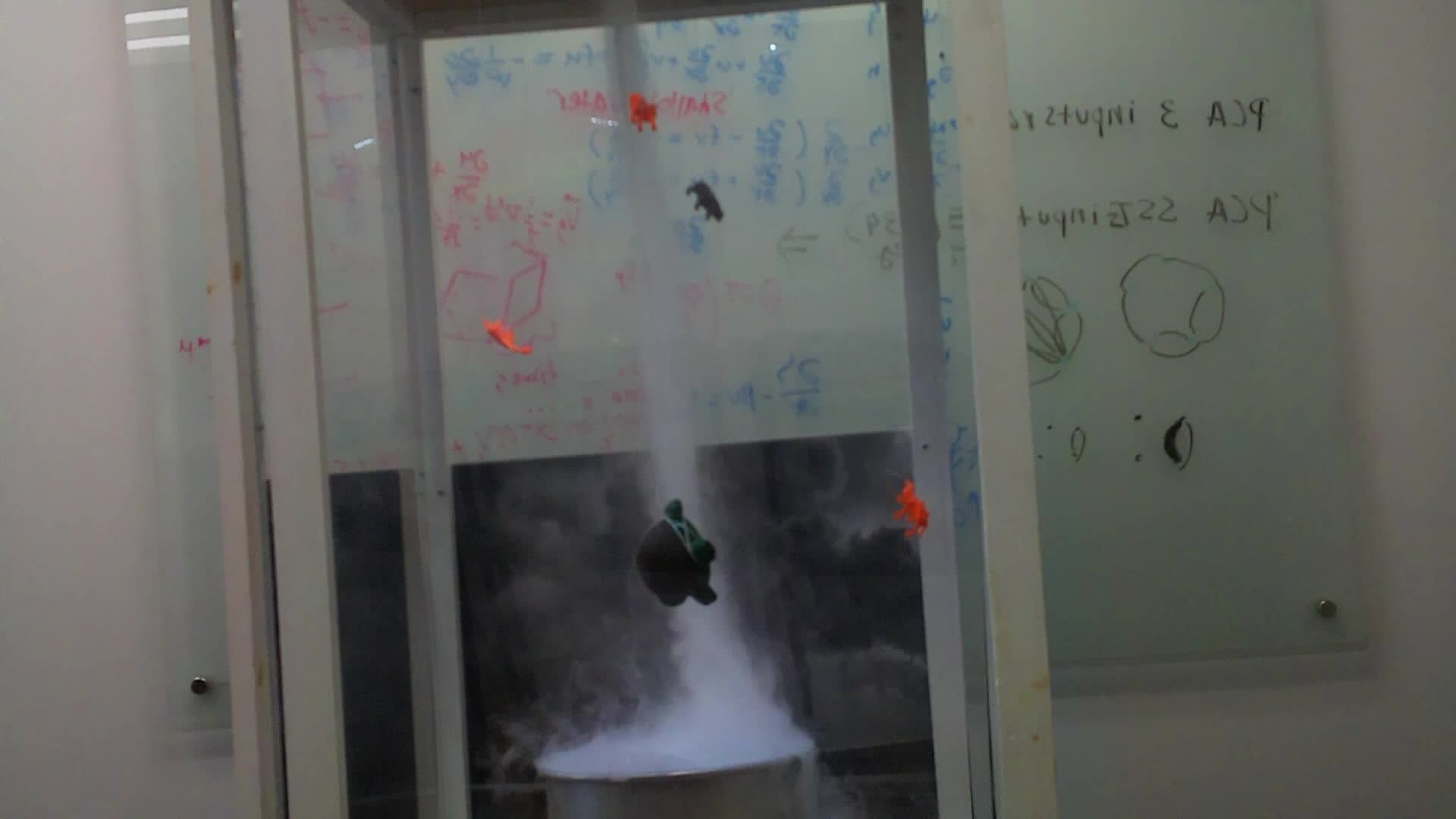} &
      \includegraphics[width=0.15\textwidth]{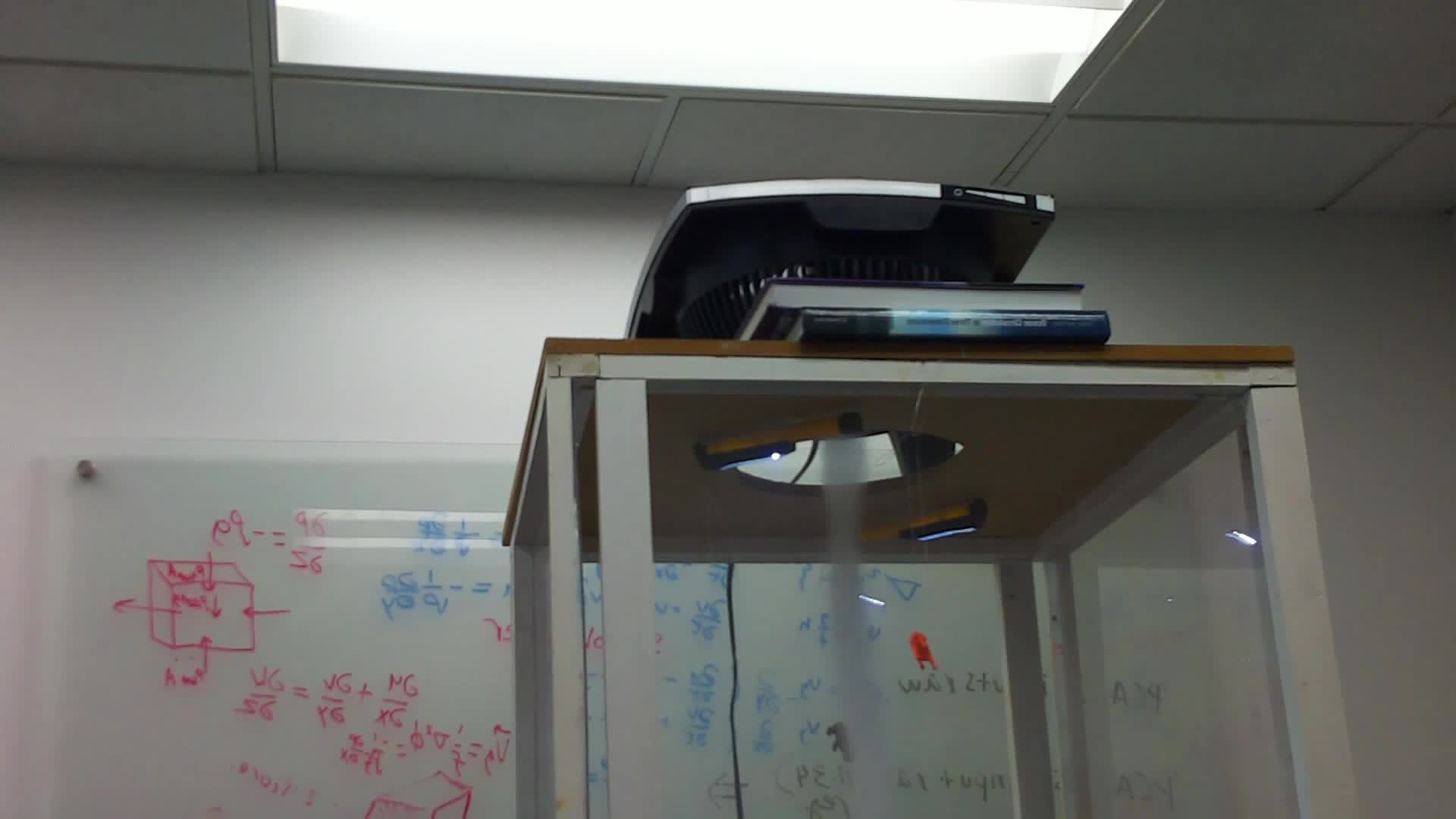} &
      \includegraphics[width=0.15\textwidth]{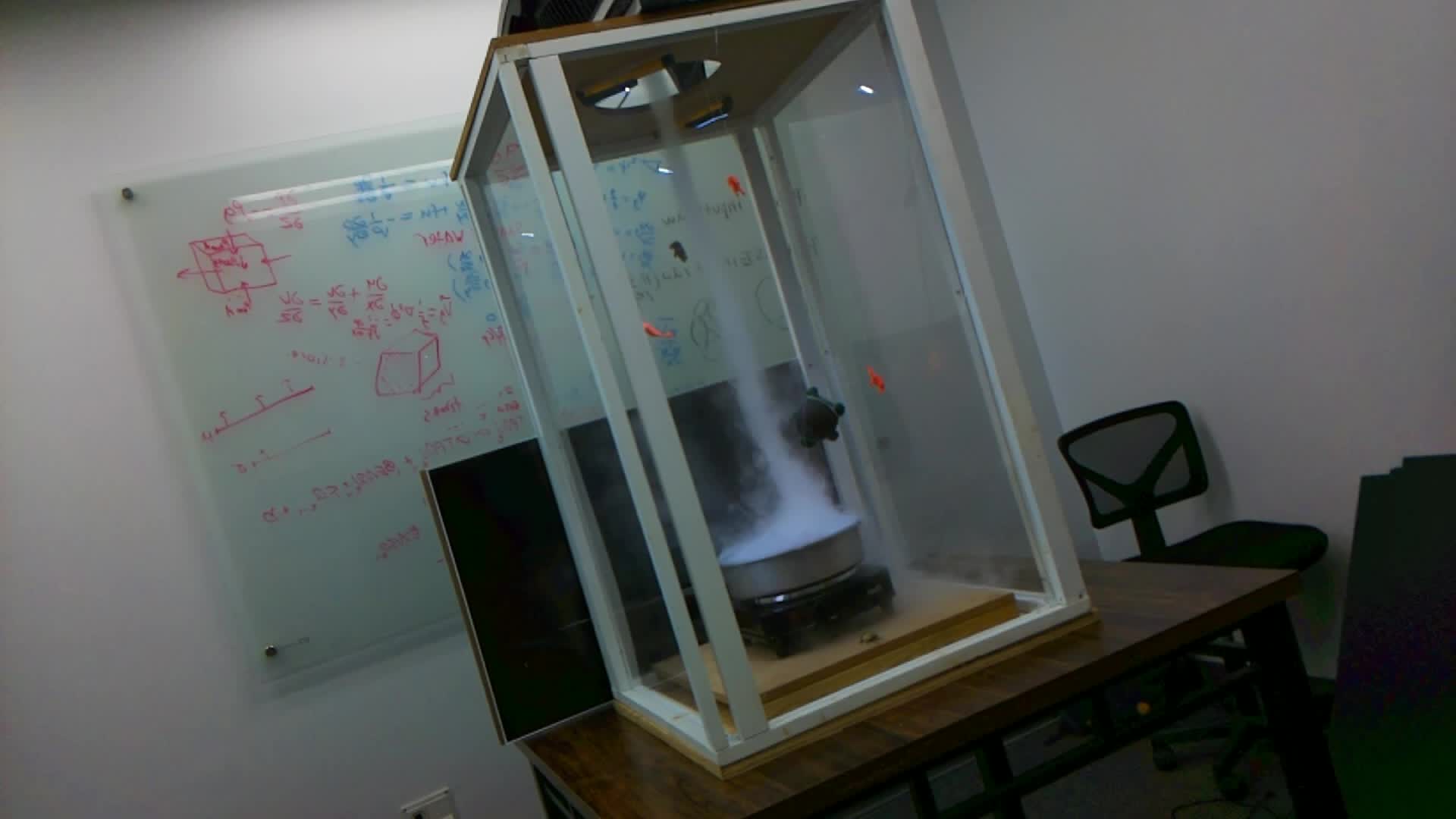}\\
      \includegraphics[width=0.15\textwidth]{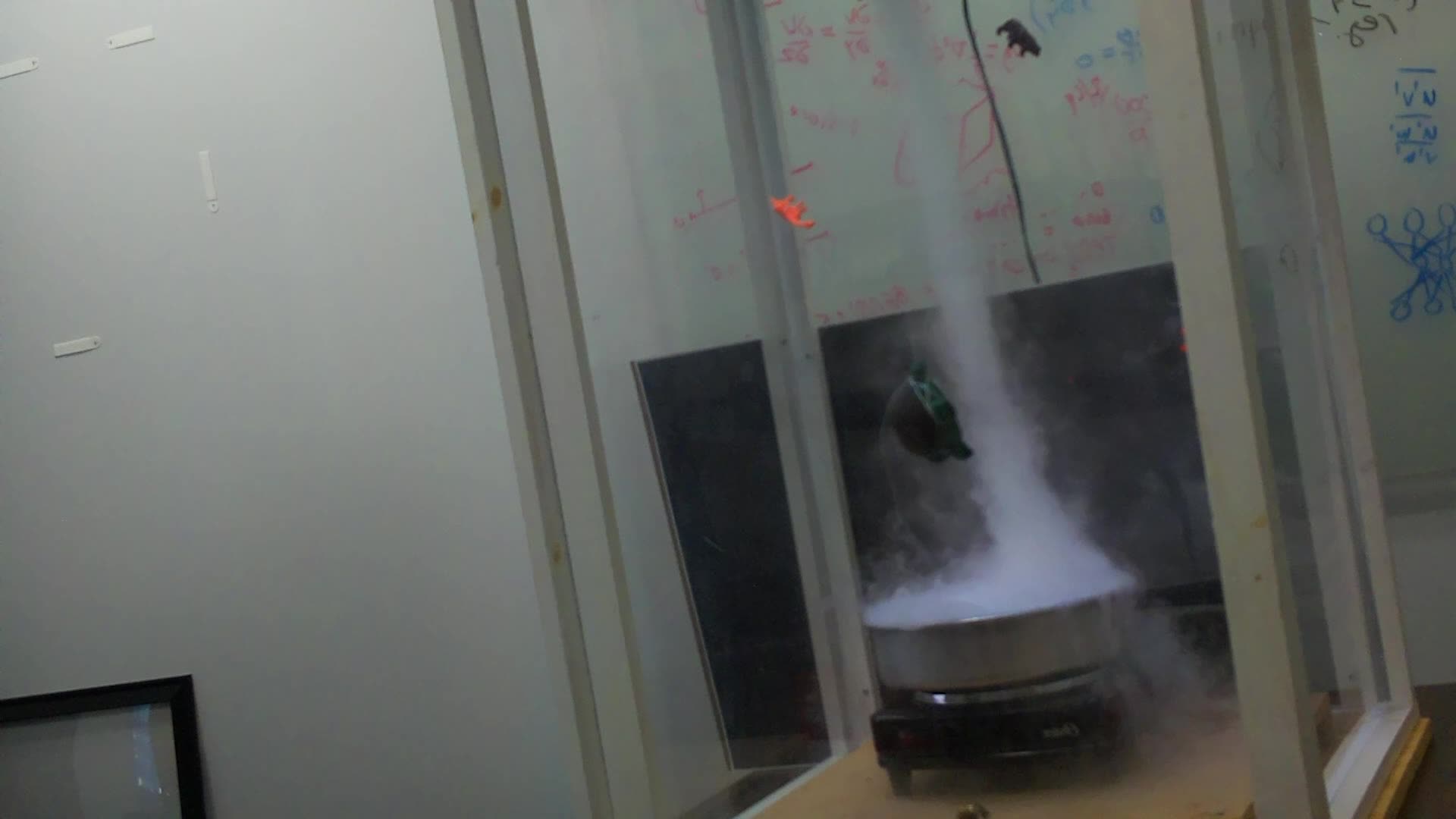} &
      \includegraphics[width=0.15\textwidth]{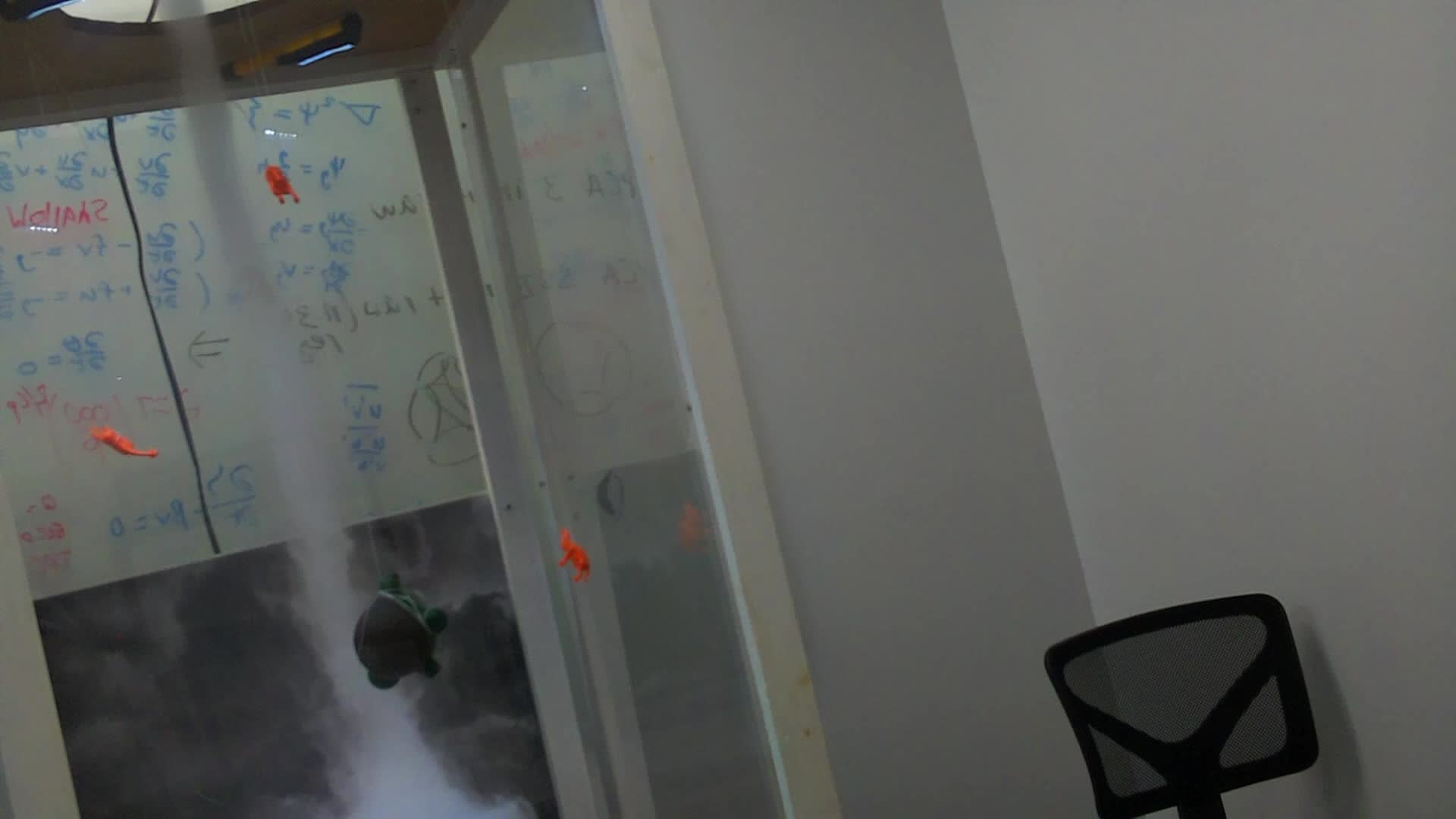} &
      \includegraphics[width=0.15\textwidth]{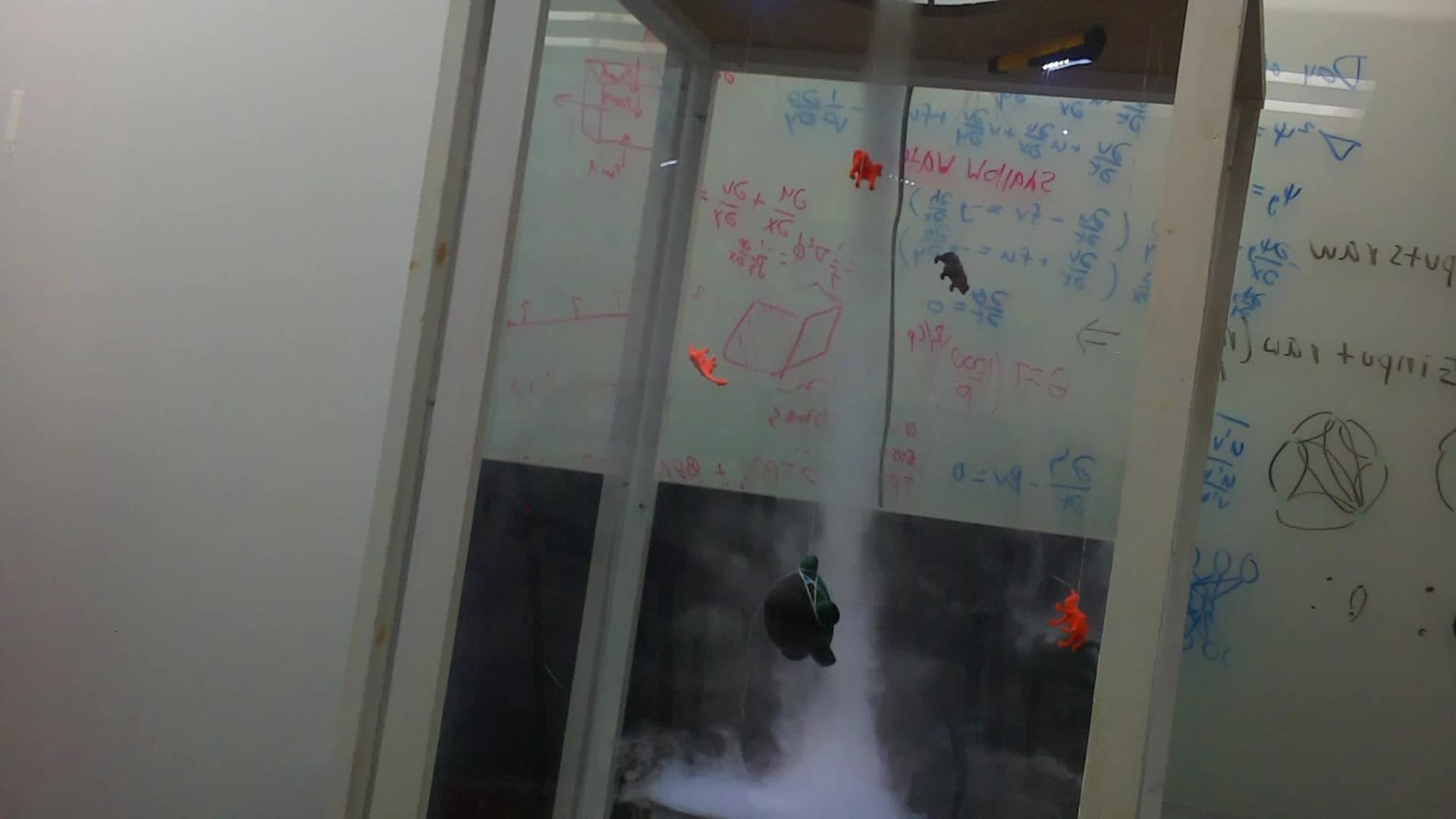} &
      \includegraphics[width=0.15\textwidth]{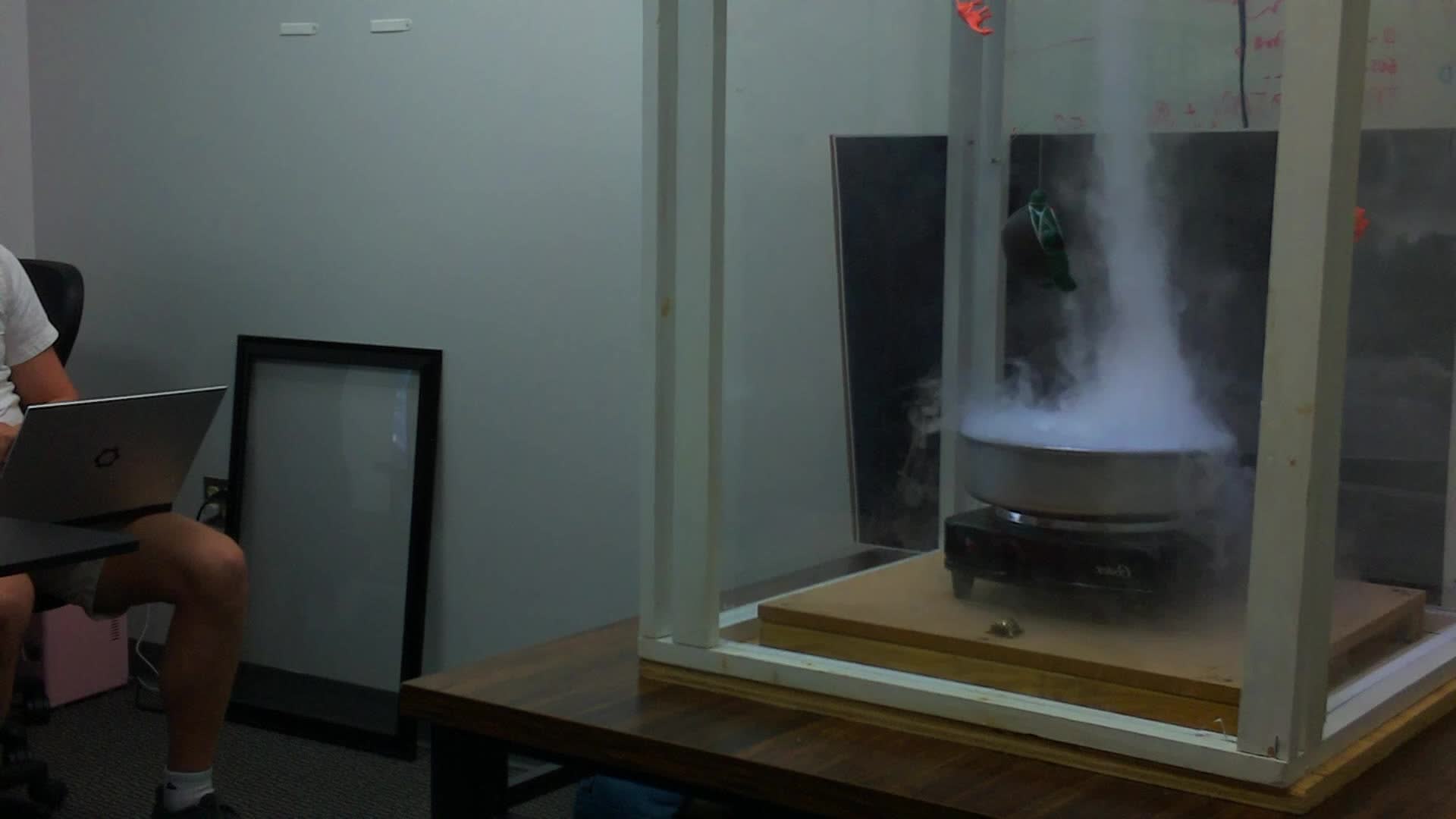} \\
    \end{tabular}
    \raisebox{-15mm}{
        \includegraphics[width=0.26\textwidth]{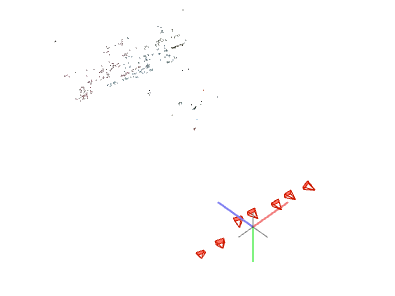}
    }
    \caption{\textbf{Multi-view setup.} Frames taken from each of the eight cameras. The images shown are used as references in the SfM pipeline discussed in Section 3.2. The resulting sparse point cloud from said images is also shown here. Note the inclusion of only seven of the eight views.}
    \label{fig:views}
\end{figure*}

\section{Background}
\subsection{Novel View Synthesis}
Novel view synthesis (NVS) is a long-standing task in computer vision that has recently been enhanced by advances in machine learning. The goal of NVS is to create unseen perspectives consistent with the 3D structure of an object of interest. Most often, this goal requires collecting multiple images of a static scene, then creating a digital reconstruction of the target to produce novel views. Neural radiance fields (NeRFs) \cite{mildenhall2020nerfrepresentingscenesneural} first introduced the idea of rendering such views using simulated light fields. Using a technique dubbed `Ray Marching,' object volume and color density can be represented using estimated light propagation to a desired (possibly unseen) camera position. While this technique can create visually compelling results, generating models can be time-consuming and computationally expensive, often requiring dense input images. This limitation motivates the application of 3D Gaussian Splatting (3DGS), proposed in \cite{kerbl3Dgaussians}. Opposed to estimating light fields, 3DGS estimates 3D structure directly using point clouds. During training, 3DGS optimizes for the density, shape, opacity, and color of deformable ellipsoids (known as Gaussians) for accurate scene reconstruction. This capability drastically reduces training time for comparably high reconstruction quality. We will deploy this technique to reconstruct a tornado.    

\subsection{Tornado NVS}
As discussed earlier, the use of NVS for the 3D visualization and modeling of atmospheric phenomena, such as tornadoes, is an underexplored topic. \cite{li2023climatenerfextremeweathersynthesis} is a NeRF-based approach for simulating real-world locations under various weather scenarios, such as flooding or snow cover. This weather scenario extrapolation is achieved by calculating the surface normals of a 3D scene and then applying visual queues to create a desired weather effect, automating the stylizing of real-world scenes. Although \cite{li2023climatenerfextremeweathersynthesis} partly deploys NVS for weather and climate tasks (i.e., simulating rare weather events), it does not adequately represent real-world phenomena, as the resulting scenes are substantial scenario extrapolations. 
\cite{AmadorHerrera:2024:Cyclogenesis} presents an NVS simulation framework for large-scale cyclones. Relying on tunable atmospheric parameters such as humidity and temperature, physical models are applied to simulate the progression of hurricanes and tornadoes. Much like the previous work \cite{AmadorHerrera:2024:Cyclogenesis}, this study introduces viable and visually compelling models of simulated scenes, rather than reconstructing existing weather phenomena.  

\section{Methods}
In the following section, we outline our progress for performing NVS on a tornado. We start by introducing the footage captured of an idealized laboratory-simulated tornado. Next, we discuss our pipeline, which applies 3DGS to the collected footage, along with the post-processing steps to enhance visual clarity. A schematic of the complete pipeline is shown in Figure \ref{fig:workflow}.

\begin{figure*}[!htb]
    \centering
    \begin{tabular}{ccc}
        \includegraphics[width=0.33\textwidth]{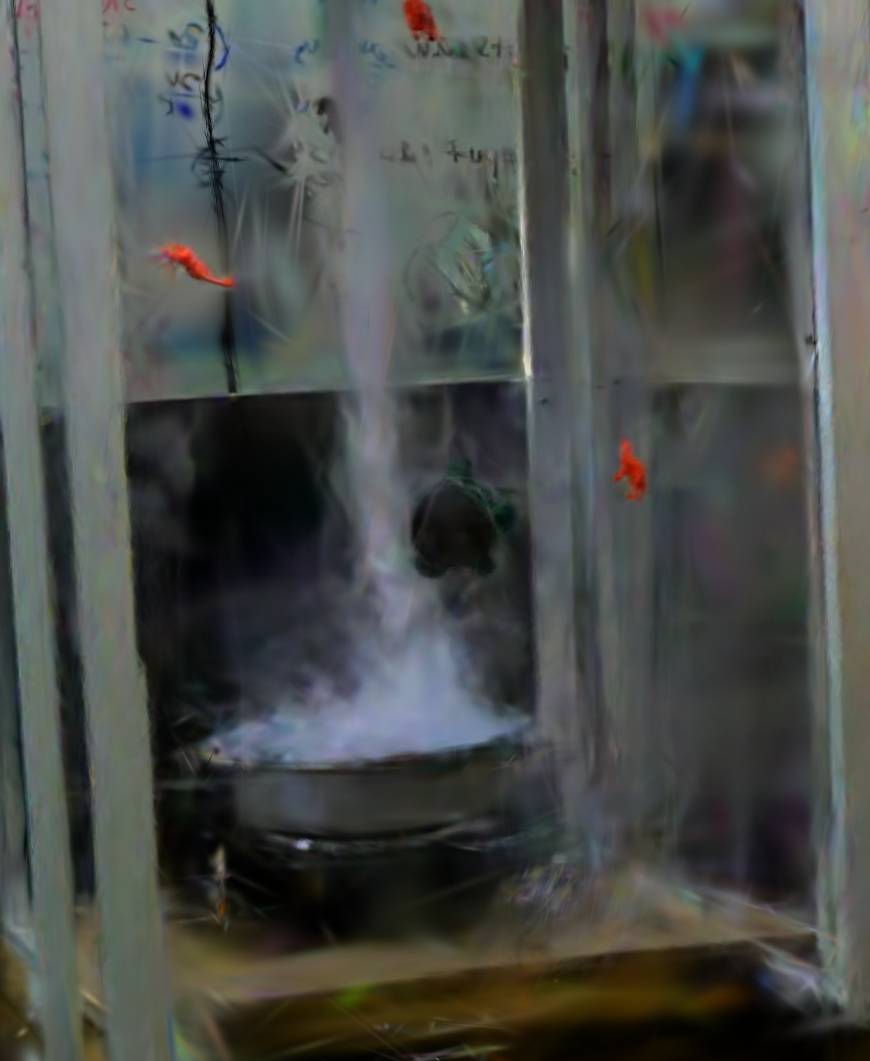} &
              \includegraphics[width=0.33\textwidth]{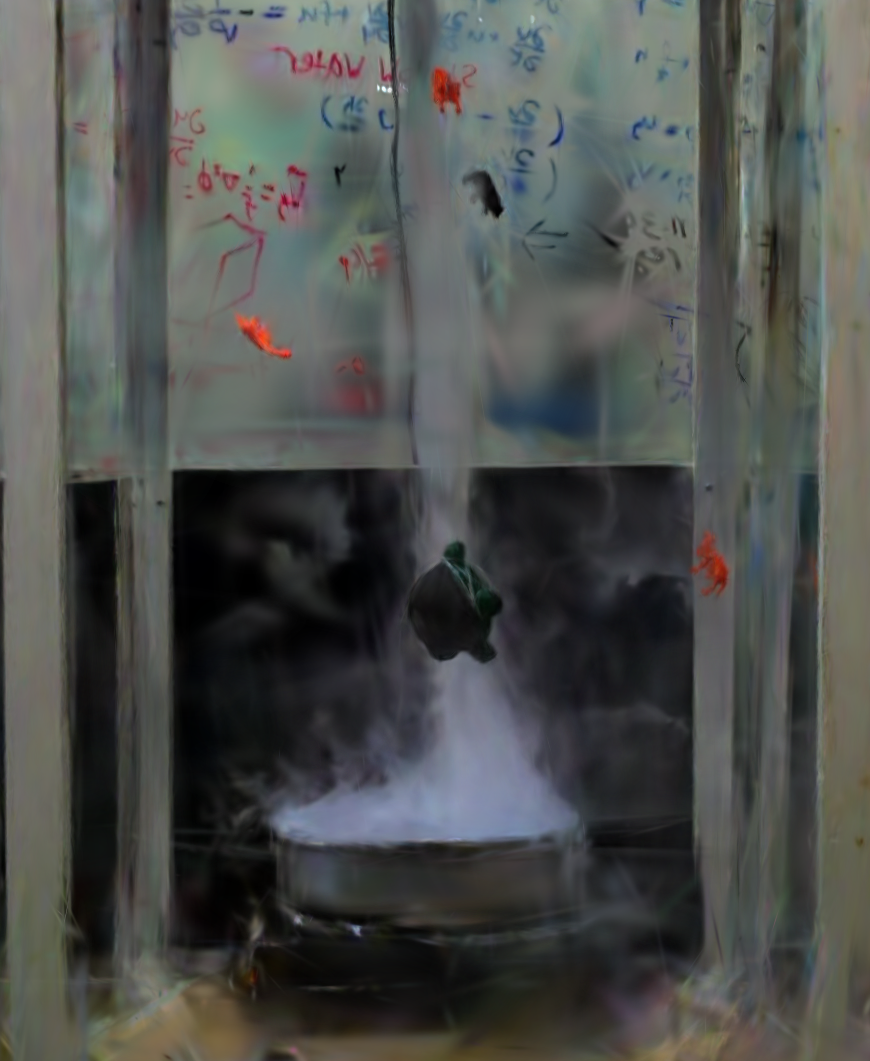} &
          \includegraphics[width=0.33\textwidth]{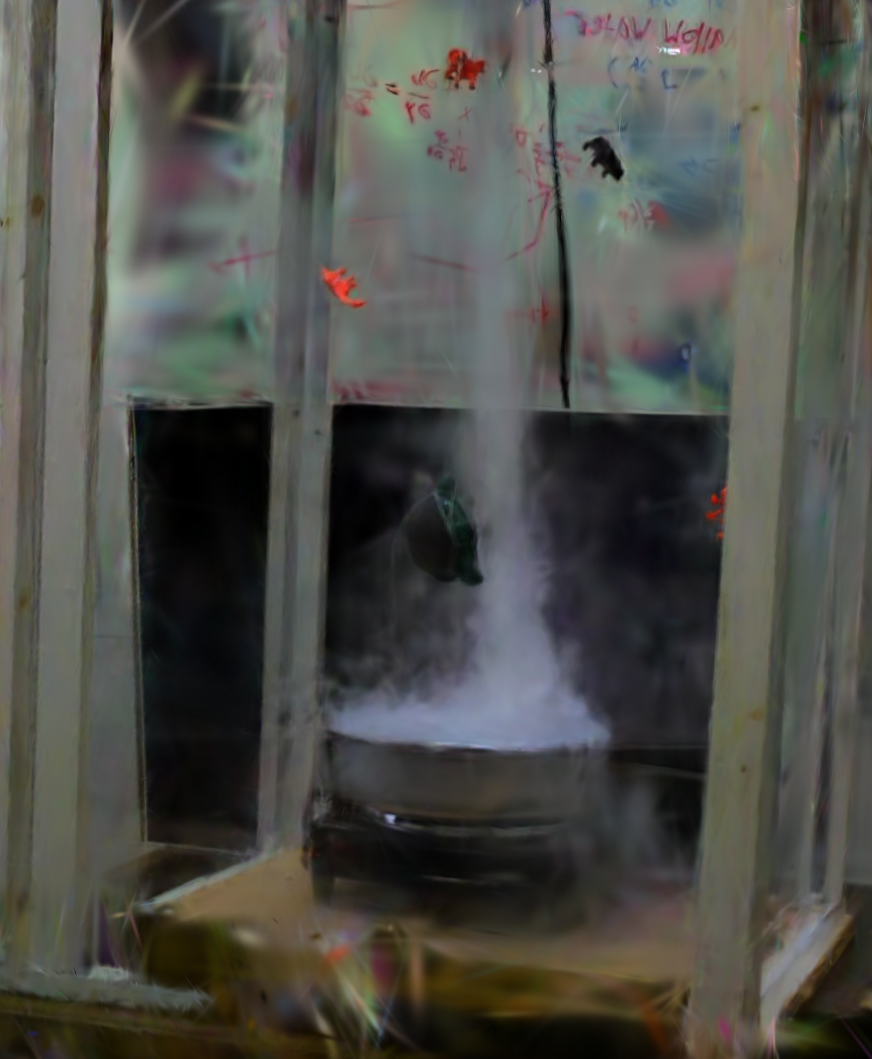} 
  \\
    \end{tabular}
    \caption{\textbf{Novel Views.} The left, middle, and right show novel views rendered from the left, middle, and right. 
    3DGS successfully recovers the 3D structure of the tornado. Please see the \href{https://adamy03.github.io/Reconstruction-Tornadoes-with-Neural-Inverse-Rendering/}{Project Page} for video fly-throughs.}
    \label{fig:output}
\end{figure*}

\subsection{Data Collection}

The obtained footage was recorded using eight Raspberry Pis, each positioned to capture different perspectives of the idealized tornado chamber as pictured. The vortex is generated within a transparent acrylic enclosure, allowing for visibility during this experiment. At each corner of the enclosure are vertical slits, angled so that air enters counterclockwise. A top-mounted fan continuously pulls air upward and out of the enclosure, drawing the angled inflow toward the center and mechanically stretching it into a vertical vortex. To mimic the role of buoyant updrafts in real thunderstorms, a heated pan of water on a raised platform provides a source of evaporating and rising warm, moist air. Dry ice is placed in the heated water pan, which sublimates and cools the surrounding water vapor, causing condensation into visible droplets that make the vortex visible. The result is pictured below (\cref{fig:views}). 

The tornado chamber provides an analogue to real atmospheric tornadoes. In nature, the counterclockwise spin of most US tornadoes originates from environmental wind shear, which creates horizontal rotation that is tilted and stretched upward by a thunderstorm’s buoyant updraft. Similarly, the tornado chamber imparts initial spin to the inflow, and when it is stretched vertically, it intensifies into a vortex, with condensation from dry ice making the circulation visible. While idealized, this setup offers a controlled environment for producing static footage. Details of the footage are outlined in \cref{tab:data_facts}.

\begin{table}[h!]
    \centering
    \begin{tabular}{|c|c|}
        \hline
        Property & Value \\
        \hline
         Frame rate & 30 \\
         Horizontal resolution & 1920 \\
         Vertical resolution & 1080 \\
         Duration & 30s \\
         \hline
     \end{tabular}
    \caption{Frame rate, resolution, and duration of captured footage. The above properties apply to each of the eight cameras.}
    \label{tab:data_facts}
\end{table}

\subsection{Model Preparation and Generation}
As discussed previously, 3DGS presents a point-based 3D rendering paradigm. The method outlined in \cite{kerbl3Dgaussians} relies on Structure from Motion (SfM) initialization produced using Colmap \cite{7780814}. Given a set of input images, Colmap estimates camera positions and points in 3D space through a process of feature extraction, matching, and camera triangulation. Using the camera views collected from the idealized tornado chamber footage, we extract frames at a common time step and estimate camera positions and intrinsics using Colmap, which is then used to initialize our 3DGS model. Our implementation follows the same workflow presented in the original 3DGS with little modification. 

\subsection{Manual Postprocessing}
While 3DGS can be used to create a rough structure of the tornado, further steps are needed for improved results. What can arise in Gaussian-based rendering techniques are `floaters,' which informally refer to errant Gaussians known not to exist in physical space. Removing these artifacts can be performed manually using a variety of rendering tools (in this work, we use \href{https://github.com/playcanvas/supersplat}{https://github.com/playcanvas/supersplat}). 

\section{Results}
In the following section, we display the results of applying 3DGS to the captured footage. Using Colmap, we obtained the following camera and point cloud model, as shown in \cref{fig:views}. Then, after training a 3DGS model on our initialization and following a postprocessing step in which Gaussians were removed for clarity, we obtained the reconstruction shown in Figure \cref{fig:output}.

\section{Discussion and Further Work}
From our results, we found a viable but flawed vanilla 3DGS pipeline to produce 3D idealized tornado models. The original reconstruction, before visual editing, included many floaters. We will discuss such errors and later suggest future directions to improve reconstruction quality.

\subsection{Camera Initialization}
One of the key challenges in 3DGS pipelines is producing adequate camera initializations. While the default SfM using colmap fails to produce an accurate camera initialization, simple adjustments (i.e., increasing the number of detected features, guided feature matching, and ignoring affine views) improve reconstruction quality. Despite this, as shown in \cref{fig:views},
our initialization only picks up seven rather than eight cameras. We anticipate that this is due to (1) high correspondence between two of the views, causing the respective view to be interpreted as one camera in the SfM pipeline, and (2) possibly due to sparse input perspectives (compared to 50-100 distinct views used in standard 3DGS pipelines). 

\subsection{Reconstruction Quality}
\paragraph{Tornado Structure} Overall, the 3D structure of the tornado was correctly produced using the default 3DGS pipeline with the addition of visual editing. For a static scene, the shape and color of the vortex were correctly produced in both original and novel views. Furthermore, other objects added for visual effect, placed inside the tornado chamber, were mostly reconstructed. 

\begin{figure}[h!]
    \centering
    \begin{tabular}{cc}
      \includegraphics[width=0.22\textwidth]{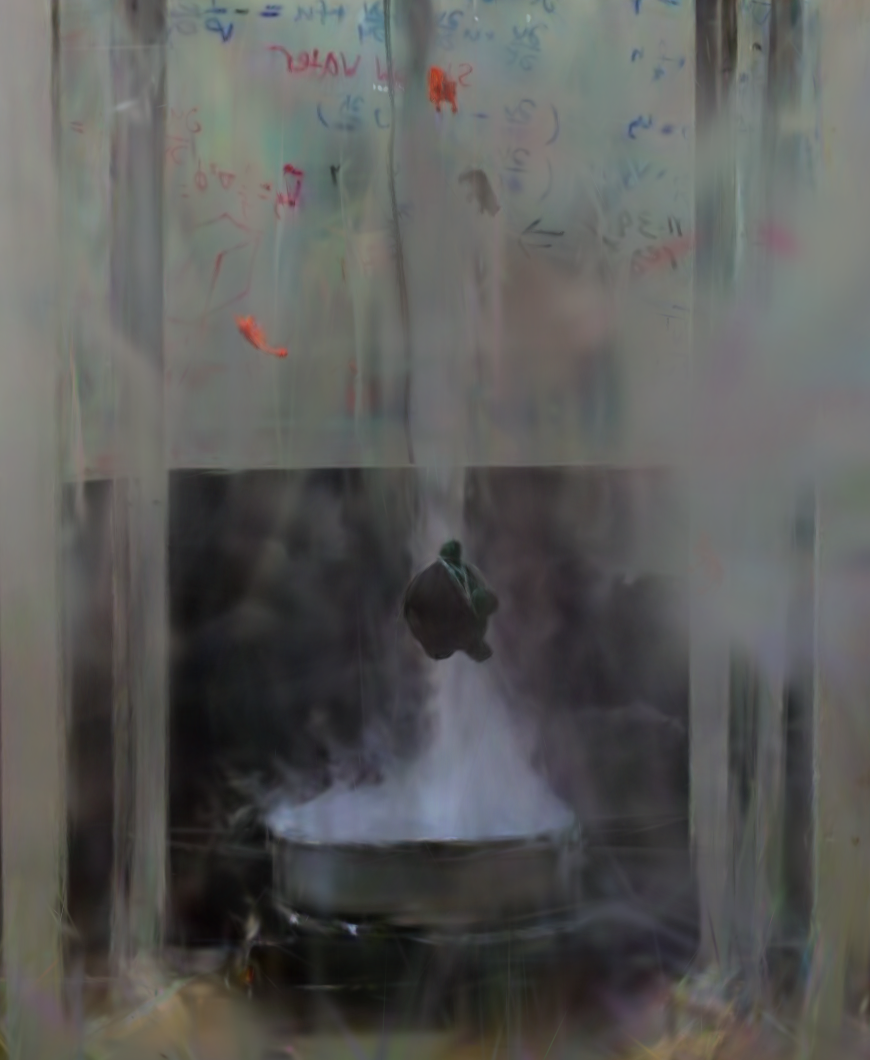} &
      \includegraphics[width=0.22\textwidth]{images/edited.png}
    \end{tabular}
    \caption{\textbf{Removing floaters.} Rendered Gaussians before and after postprocessing. The left image shows the 3DGS model before postprocessing, and the right image shows the postprocessed model. Floaters show up in novel views, but can be reduced by using more accurate camera initializations, optimizing 3DGS hyperparameters, and a separate pruning post-processing step.}
    \label{fig:output_manual}
\end{figure}

\paragraph{Manual Postprocessing} 
As discussed earlier, the resulting reconstruction required further editing to create visually compelling results. Figure \ref{fig:output_manual} illustrates the impact of postprocessing.

Novel views produced from reconstructions before postprocessing are occluded by `floater' Gaussians, which must be removed. Such artifacts are not consistent with the overall 3D structure and are a byproduct of the 3DGS optimization process. This may be improved by employing several methods beyond simply adding more perspectives to enhance camera views. Later works can explore alternatives for point cloud initialization (such as random initialization \cite{jung2024relaxingaccurateinitializationconstraint} \cite{foroutan2024evaluatingalternativessfmpoint}) or skipping the SfM process entirely \cite{fu2024colmapfree3dGaussiansplatting}. 

\section{Conclusion}
In this work, we demonstrate the feasibility of Gaussian splatting for NVS of a tornado in a controlled environment. Using vanilla 3DGS and manual postprocessing, a viable structure is produced using eight distinct views. While imperfect, this exemplifies the promise of standard NVS algorithms, aided by manual postprocessing, for reconstructing a relatively featureless target object. Our work presents preliminary steps for conducting NVS on a small, idealized lab-based tornado; several steps remain before this methodology can be performed in the wild.

Tornadic activity has been shifting towards the southeastern US in relation to climate change \cite{gensini_spatial_2018}, where safe observation of tornadoes is increasingly difficult and tornado-related risks are higher than in the Great Plains \cite{strader2024changes}. Understanding the 3D wind field of tornadoes to advance knowledge and improve forecasting may require utilizing existing datasets in novel ways.

\subsection*{Acknowledgments}
The authors thank Anton Seimon, whose amazing storm chasing footage helped inspire this work. We also acknowledge the undergraduate students in the Department of Atmospheric and Oceanic Science at the University of Maryland, who built the tornado generator in 2013.

{\small
\bibliographystyle{ieee_fullname}
\bibliography{egbib}
}

\end{document}